%% file: charac.tex
\documentclass[conference]{IEEEtran}
\usepackage{cite}

\usepackage{tikz,pgf,pgfplots}
\pgfplotsset{compat=1.14}
\usepackage{verbatim}
\usepackage{array, booktabs, makecell}
\usepackage{dirtytalk}
\usepackage{siunitx}
\usepackage{epsfig}
\usepackage{subfigure}
\usepackage{graphicx}

\usepackage{url}

\usepackage[utf8x]{inputenc}
\usepackage{amsmath,amsthm, amsfonts,amssymb}
\usepackage{mathtools}

\usepackage{color}

\begin{document}

\title{Characterising Across-Stack Optimisations for \\ Deep Convolutional Neural Networks}

\author{\IEEEauthorblockN{Jack Turner,
Jos\'e Cano,
Valentin Radu,
Elliot J. Crowley, 
Michael O'Boyle,
Amos Storkey}
\IEEEauthorblockA{
       \begin{tabular}{ccc}
               {School of Informatics}
       \end{tabular} \\
				\begin{tabular}{ccc}
               {University of Edinburgh, UK}
       \end{tabular}
   }		
}

\maketitle

\begin{abstract}

Convolutional Neural Networks (CNNs) are extremely computationally demanding, presenting a large barrier to their deployment on resource-constrained devices. Since such systems are where some of their most useful applications lie (e.g.\ obstacle detection for mobile robots, vision-based medical assistive technology), significant bodies of work from both machine learning and systems communities have attempted to provide optimisations that will make CNNs available to edge devices. In this paper we unify the two viewpoints in a Deep Learning Inference Stack and take an across-stack approach by implementing and evaluating the most common neural network compression techniques (weight pruning, channel pruning, and quantisation) and optimising their parallel execution with a range of programming approaches (OpenMP, OpenCL) and hardware architectures (CPU, GPU). We provide comprehensive Pareto curves to instruct trade-offs under constraints of accuracy, execution time, and memory space.
\end{abstract}

\IEEEpeerreviewmaketitle

\input{introduction}

\input{background}

\input{experiments}

\input{conclusions}

\section*{Acknowledgment}

This project has received funding from the European Union's Horizon 2020 research and innovation programme under grant agreement No 732204 (Bonseyes). This work is supported by the Swiss State Secretariat for Education, Research and Innovation (SERI) under contract number 16.0159. The opinions expressed and arguments employed herein do not necessarily reflect the official views of these funding bodies. The authors are grateful to Lizhong Chen and the anonymous reviewers for their valuable contributions.

\bibliographystyle{IEEEtran}
\bibliography{charac}

\end{document}

%% file: introduction.tex
\section{Introduction}

Recent years have yielded rapid advances in the field of deep learning, largely due to the unparalleled effectiveness of Convolutional Neural Networks (CNNs) on a variety of difficult problems~\cite{lecun2015deep}. These networks are designed to run on servers with 
negligible resource constraints, 
utilising powerful GPUs. As such, creative approaches are required to deploy them on hardware with limited resources in order to enable many useful applications (e.g.\ autonomous driving~\cite{bojarski16,teichmann16}, collision avoidance for quadcopters~\cite{alvarez16}, human activity recognition with wearable sensors~\cite{radu18}, medical systems~\cite{doherty13}) building on CNN-based detections. However, currently these optimisation approaches come with limited benchmarks and few comparisons. We outline a first step towards a more comprehensive understanding of the performance available under different constraints of inference accuracy, execution time, and memory space.

Since~\cite{krizhevsky2012imagenet} used CNNs to outperform more traditional statistical techniques on the ImageNet dataset~\cite{ILSVRC15} they have become a standard tool for image processing. With a growing ecosystem dedicated to training deep neural networks, the number of parameters that state-of-the-art networks demand has vastly increased; in 2012 the state-of-the-art, AlexNet, had 61M parameters spread over eight layers whereas the most recent ImageNet winner uses an ensemble of SENets~\cite{SEnet}, the largest of which has 115M parameters across 154 layers.

\begin{figure}[t]
\includegraphics[width=\linewidth]{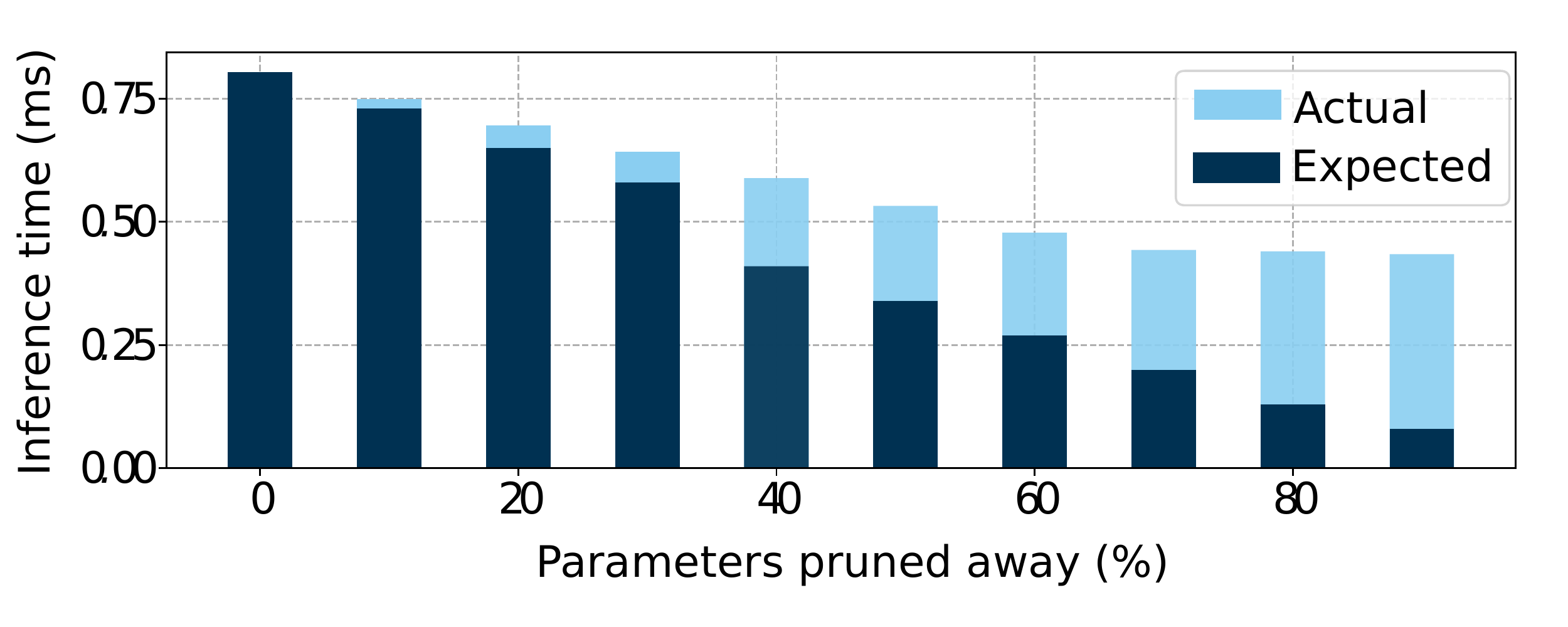}
\vspace{-0.5cm}
\caption{Expected vs. observed inference time for VGG-16 on an Intel Core i7 processor when compressed with a popular weight pruning scheme~\cite{han2015learning}.}

\label{fig:expected-inf-time}
\vspace{-0.3cm}
\end{figure}

This large number of parameters comes with a significant computational burden reflected in memory, compute time, and energy consumption. However, there is a great deal of redundancy in these parameters; in~\cite{denil2013predicting} the authors demonstrate that up to 95\% of the weights\footnote{The parameters in neural networks are commonly referred to as weights, and as such, we use these terms interchangeably.} in a CNN can be predicted from a small subset of weights without affecting accuracy. This is particularly prudent given that~\cite{han2015deep} showed that the bottleneck for inference computation was off-chip DRAM accesses, and that when the memory requirements of a CNN are reduced, the energy consumption and the time taken to compute an inference are also reduced. 

In light of this, a range of compression techniques have been developed to reduce the number of parameters in CNNs~\cite{han2015deep,ye2018rethinking,dong2017learning,louizos2017bayesian,moonshine,condensenet}. Alternatively, networks can be compressed through quantisation, for example by reducing the number of bits used to represent each parameter~\cite{zhou_2017_incremental} or by constraining their possible values~\cite{courbariaux_2015_binaryconnect,chen_2015_hashing}. Despite the impressive experimental results of these techniques, they are rarely paired with hardware performance characterisations. In fact,~\cite{yu2017scalpel} showed that some types of compression can \textit{hurt} the hardware performance of these networks despite significant reductions in multiply-accumulate (MAC) operations; even where inference time improvements did exist, they did not align with the expected speedup. We verify this with a motivating example presented in Figure \ref{fig:expected-inf-time}, which shows the gap between expected performance (calculated based on the number of operations that need to be performed) and observed performance for increasing compression rates on VGG-16, a popular CNN. 

As a consequence, deep learning researchers and practitioners are left to explore a vast space of possible optimisations with poor intuition as to the effect they will have on real hardware performance. We believe this issue exists primarily due to a fundamental separation between machine learning experts and system design engineers, each taking an independent view of their domain-specific problems, which is limiting the advancement of both fields. We aim to bridge this gap by taking a bi-directional view, both from a machine learning perspective and a system level perspective to unify observations, which are useful to instruct both sides.

Inspired by~\cite{yu2017scalpel}, we explore a broad range of compression techniques emerging from the machine learning domain. Specifically, we evaluate parameter pruning, channel pruning and quantisation on different CNN models (VGG-16~\cite{vgg}, ResNet-18~\cite{he2016deep} and MobileNet~\cite{mobilenet}) adapted for image classification on the CIFAR-10~\cite{cifar} dataset. As system level optimisations, we employ commonly-used libraries to exploit the parallelism options present on resource-constrained devices. Execution time, memory footprint, and inference accuracy are the primary metrics used in this work to evaluate the impact of the explored solutions. This paper makes the following contributions:

\begin{itemize}
\item We introduce the concept of a \textit{Deep Learning Inference Stack} to reflect the different layers\footnote{These layers are not to be confused with the computational layers that make up a neural network.} of optimisation where candidate solutions can be applied to make a neural network model run more efficiently on resource-constrained devices.

\item At each layer of this stack we identify the most relevant candidates from both the machine learning community (CNN topologies, compression techniques), and the systems community (code optimisation and parallelism) for accelerating neural networks.

\item Candidate solutions are implemented and evaluated across the layers of the stack to produce across-stack observations and characterise their impact. Our network implementations are available online\footnote{https://github.com/jack-willturner/characterising-neural-compression} for the community to expand on our work.

\item We provide guidelines for adapting established neural network architectures to run on resource-constrained hardware. Given constraints of accuracy, inference time, and memory footprint, we demonstrate that significant performance enhancements can be achieved given insights from our exploratory work. We show that compression techniques can be applied to large networks to surpass the performance (inference time with fixed accuracy) of smaller networks that were handcrafted for embedded devices, e.g.\ compressed VGG-16 vs. MobileNet.

\end{itemize}

The rest of the paper is organised as follows: In Section~\ref{sec:stack} we describe the full stack used at both the machine learning and system levels for this characterisation. In Section~\ref{sec:background} we cover techniques in the machine learning literature used to reduce the computational complexity of neural networks. This is followed by a presentation of our experimental setup and choice of solutions (Section~\ref{sec:setup}). In Section~\ref{sec:experiments} we provide our experimental results and recommendations for deploying neural networks on resource-constrained devices. Finally, in Section~\ref{sec:discussion} we discuss the results of our experiments.

%% file: background.tex
\section{The Deep Learning Inference Stack}
\label{sec:stack}

\subsection{Motivation}

The recent growth of deep learning has been facilitated by the availability of massive computational power on clusters of computers. When combined with a tendency to focus narrowly on inference accuracy, this has led to state-of-the-art CNNs exploding in size. This presents a large barrier to deploying many modern deep learning applications on embedded devices.

Both machine learning researchers and system design engineers have proposed innovative solutions to overcome this barrier. However, these solutions are typically developed in isolation. For instance, sparsity is regarded by some in the neural network community as a silver bullet for compressing models, whereas exploiting parallelism is generally seen as essential for neural network computations by system architects. Challenging these isolated preconceptions reveals that sparsity does not always excel at reducing the number of operations during inference, and parallelism does not necessarily come with the speedups expected on neural network workloads. These observations are presented in greater detail in Section~\ref{sec:experiments}. 

It is clear that we need a better approach to assess the proposed techniques and optimisation solutions that range across these two disciplines. With this paper we take a first step in providing this comprehensive analysis by exploring a range of machine learning and system level techniques which shed light on the current state of development across the two disciplines targeting resource-constrained devices. We hope that our methodological approach to join the two disciplines will lead to further common efforts to advance the fields.

\begin{table}[t]
\begin{center}
\vspace{-0.3cm}
\caption{The Deep Learning Inference Stack.}
\begin{tabular}{ |c|c|c| } \hline
\bf{\#}&\bf{Layer Name} & \bf{Short Description} \\ \hline \hline
1&Neural Network Models & Dense neural network models \\ \hline
2&Machine Learning Techniques & Common compression techniques \\ \hline
3&Data Formats and Algorithms & Network weights representation \\ \hline
4&Systems Techniques & Parallelisation, optimisations \\ \hline
5&Hardware & Resource-constrained devices \\ \hline
\end{tabular}
\label{table:inference_stack}
\end{center}
\end{table}

\subsection{Description of the Inference Stack}

We introduce the \textit{Deep Learning Inference Stack}, which spans from the machine learning domain all the way down to the hardware domain. Each layer can be tuned to optimise different goals (i.e.\ inference accuracy, execution time, memory footprint), or to yield further improvements in adjacent layers. The Stack contains the following layers (Table~\ref{table:inference_stack}):

\begin{enumerate}

\item \textit{Neural Network Models} -- This encompasses the full-precision dense models proposed in the machine learning community designed for particular tasks (e.g.\ image classification, speech recognition).  

\item \textit{Machine Learning Techniques} -- The choice of compression techniques, such as pruning or quantisation.

\item \textit{Data formats and Algorithms} -- Selecting the right data format for each compression technique is important to avoid penalties imposed by memory access requirements (e.g.\ speed, bandwidth) and to improve memory locality. Interlinked with data formats are computation algorithms (e.g.\ direct convolutions, image2col) and other data transformations (e.g.\ Winograd transform).

\item \textit{Systems Techniques} -- Inference algorithms in neural networks are ideal for parallelisation, although their efficiency is specific to the software libraries, programming languages, and compilers used. 

\item \textit{Hardware} -- This layer includes all resource-constrained devices these models can be run on, which are fundamentally different to the servers where the original models were designed to run (GPUs in clusters).

\end{enumerate}

Design decisions made at each layer of the stack directly impact adjacent layers. They can also influence decisions across the entire stack. In this paper we perform an across-stack investigation of the different techniques at each layer to determine their impact on inference and runtime performance. By bridging together concepts from both machine learning and systems domains in one accessible stack, we can expose limiting preconceptions in each domain and identify their impact on the whole stack.

%---------------------------------------------------------------------------------------------------------------------------------------------------------

\section{Background}
\label{sec:background}

\begin{figure*}[!ht]
%\vspace{-0.25in}
\begin{center}
\begin{tabular}{cc}
\subfigure[Dense Convolutions]
{
  	\includegraphics[width=0.4\textwidth]{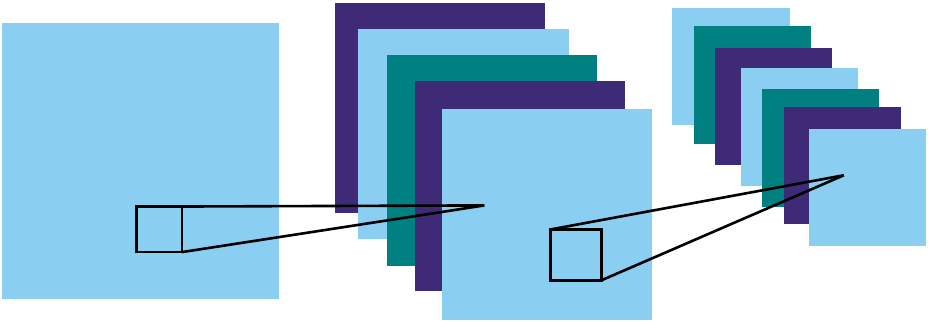}
	\vspace{2cm}
\label{fig_dense}
}&

\subfigure[Weights Pruning]
{
  \includegraphics[width=0.4\textwidth]{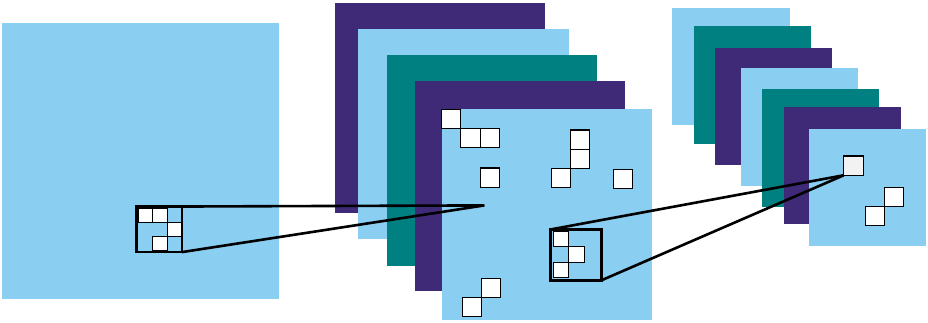}
  \label{fig_sparse}
}
\\
\subfigure[Channel Pruning]
{
  \includegraphics[width=0.4\textwidth]{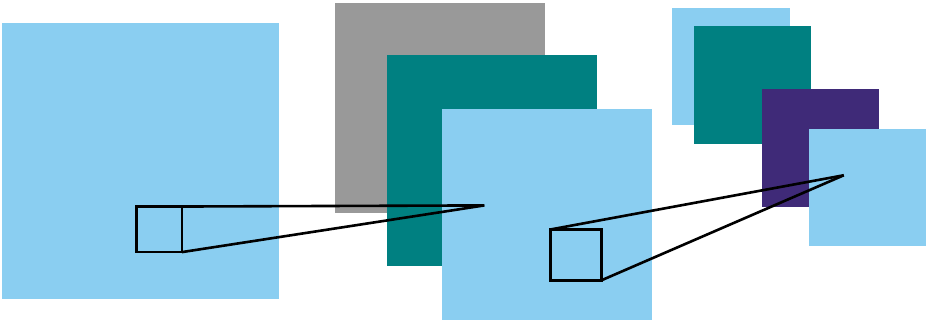}
  \label{fig_fisher}
}&
\subfigure[Weights Quantisation]
{
  \includegraphics[width=0.18\textwidth]{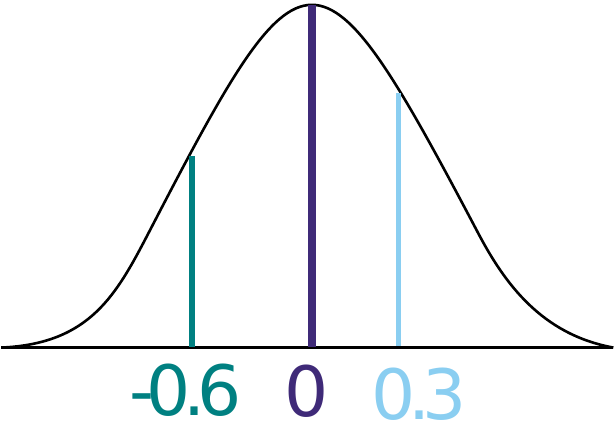}
  \vspace{5cm}
  \label{fig_quantization}
}
\\
\end{tabular}
\caption{A visual representation of different compression techniques: (a) shows a slice of a standard neural network. Each rectangle represents the output generated by convolving filters with learnt weights (parameters) over each input channel; (b) shows the same network slice with weight pruning applied. A subset of weights in the filters are forced to zero (visually represented with white holes) producing sparse matrices; (c) shows the network slice with channel pruning applied onto the middle layer, where the connections remain dense but there are fewer channels; (d) shows the quantisation process for weights, constraining all value from the original distribution to just three possible values (ternary quantisation).} 
\label{fig:systems}
\end{center}
\end{figure*}

A range of compression techniques have been proposed in the Machine Learning community. In this section we present the techniques most relevant to this paper and their variants.

\subsection{Weight pruning} % Jack

Weight pruning is a popular technique for reducing the parameter count in neural networks, and has been shown to improve generalisation~\cite{scardapane2017group}. Weight pruning techniques introduce sparsity in the network and generally fall into two categories: (i) unstructured removal of weights deemed unnecessary; and (ii) structured regularisation --- pushing the network weights towards a regular sparse format. The step from Figure \ref{fig_dense} to \ref{fig_sparse} shows an example of a network with individually pruned weights. 

The removal of redundant weights was first proposed in~\cite{NIPS1988_156}, which uses different biases to penalise the error function used to train the network. A more computationally demanding approach was suggested in~\cite{lecun1990optimal} which involved computing second derivatives to find and remove weights with the smallest effect on the error function.  

A less computationally demanding method is proposed in~\cite{han2015learning}, where the network is trained in its original dense format, and then all weights below a certain threshold are removed. This is applied layer-by-layer, where the threshold is determined by the standard deviation of the layer. The authors show that after pruning, the network should be retrained with its new weights in order to maintain accuracy, and that by iteratively repeating this process they can prune up to 90\% of the weights without affecting accuracy. In~\cite{han2015deep}, this is developed using a three stage method for storing the network involving pruning, quantisation, and Huffman coding.

Other works \cite{liu2015sparse,denil2013predicting} have utilised low-rank approximations that yield significantly reduced sparse networks. The major benefit of this approach is that the network can be kept in a highly optimisable form. In~\cite{liu2015sparse} the authors present a sparse convolution algorithm that can be used along with compression methods to outperform the original dense network. Similarly, \cite{scardapane2017group,wen2016learning} use sparse group Lasso to penalise small weights and push the network towards a structured sparse format.

\subsection{Channel pruning} 

A natural extension of weight pruning is to not only remove individual weights in a network, but entire channels from the convolutional layers and nodes from the fully connected layers, as shown in Figure \ref{fig_fisher}. One approach to this is to apply Lasso regression to the problem of channel selection~\cite{he2017channel}. This view of channel pruning was questioned by~\cite{ye2018rethinking}, who applied Lasso instead to the batch-norm~\cite{ioffe2015batch} layers of the network and then removed any channels with low batch-norm weights, resulting in greater accuracy. 

Another approach altogether is to approximate the effect of removing certain channels on the error using a \emph{Taylor series} expansion~\cite{molchanov2016pruning,theis2018faster}. Surprisingly, it has been shown that random pruning~\cite{mittal2018recovering} is also an effective strategy for removing filters; many common networks are able to retrain to their original accuracy after iterative stages of randomly removing progressively more filters.

\subsection{Quantisation} 

Quantisation methods attract a lot of attention in the machine learning community due to their immediate impact on memory requirements. In~\cite{zhou_2017_incremental} the number of bits used to represent each weight is reduced. Others have instead restricted each weight to one of several fixed values. HashedNet~\cite{chen_2015_hashing} restricts weights to a smaller set of possible values by using a hash function to map weights to hash buckets, in which they share the same floating point value. The extreme case is achieved by BinaryNet~\cite{courbariaux_2015_binaryconnect} transforming all weights to a one bit representation, with minimal accuracy degradation. 

These quantisation methods have been proposed and evaluated predominantly on well-known CNNs starting from a floating-point weight representation. The networks are typically pre-trained and then quantisation is applied gradually while fine-tuning (i.e.\ re-training) to maintain a high inference accuracy. This fine-tuning process is essential to compensate for the loss of information in quantising weights. 

Several works observe an increase in network performance after this process, explainable by quantisation acting as a regularisation method in the fine-tuning phase, nudging the quantised model in the right direction in a reshaped convex space. One such promising approach that combines elements from previous quantisation methods is presented in~\cite{zhu_2017_trained}. This ternary quantisation method transforms the weights in convolution filters to just three possible values, the zero value and two other values (one positive and one negative) determined independently for each layer. Figure \ref{fig_quantization} shows an example. %:

%% file: experiments.tex
\section{Experiments Setup}
\label{sec:setup}

%------------------------------------------------------------------------------------------------------------------------------------------------------

We evaluated each network model on the CIFAR-10 dataset~\cite{cifar}, which is a standard benchmark in machine learning and computer vision. It comprises of 60,000 images across 10 object categories, split into a training and test set of 50,000 and 10,000 images respectively. The size of each image is $32\times32$ pixels represented in RGB format. During training we augment the data by padding each image with $2\times2$ zeros and taking random $32\times32$ crops.

\subsection{Neural Networks Models} 
\label{sec:hyperparameters}

We consider the workloads of three CNNs that represent three separate classes of network topology. To train the models we used Stochastic Gradient Descent (SGD) to minimise the cross-entropy loss (averaged across all data items), which penalises the network for making incorrect classifications. We use a stepped learning rate, starting at $0.1$ and decreasing by a factor of $10$ every 50 epochs. 

The first network, VGG-16~\cite{vgg}, is a feed-forward network with 13 convolutional layers and 3 fully connected layers. Each convolution uses a $3\times3$ kernel. There are max-pooling operations after layers~\{2, 4, 7, 10, 13\}. Since VGG-16 is defined for the ImageNet dataset, we use a truncated CIFAR-10 form that replaces the final three fully connected layers with two layers containing 512 and 10 nodes respectively.

ResNet-18~\cite{he2016deep} is an 18 layer convolutional network that consists of residual blocks. Each block contains two convolutional layers (using $3\times3$ kernels) and blocks are connected in a feed-forward manner. Additionally, there is a \textit{skip connection} between the input and output of each block, which helps gradients propagate through the network~\cite{shatteredgradients2017}. This network also utilises batch normalisation~\cite{ioffe2015batch} after each convolution and is defined for CIFAR-10.

The final network we consider is MobileNet~\cite{mobilenet}, which was developed specifically for embedded devices. It takes advantage of \textit{depthwise separable convolutions} in place of the traditional convolution operation, allowing for a significant reduction in computation and parameter costs. MobileNet consists of 27 convolutional layers, alternating between $3\times3$ depthwise convolutions and $1\times1$ pointwise convolutions, and a single fully connected layer for classification. We use the original ImageNet definition of MobileNet but change the number of outputs from 1,000 to 10 to match the number of object categories present in CIFAR-10.

%------------------------------------------------------------------------------------------------------------------------------------------------------

\subsection{Machine Learning Techniques}

In our characterisation we use Deep Compression~\cite{han2015deep}, which is the state-of-the-art for weight pruning. To represent channel pruning we use Fisher pruning~\cite{theis2018faster,molchanov2016pruning}, as it yields the most impressive levels of compression for this category. Finally, we use the ternary quantisation method~\cite{zhu_2017_trained} because: (i) it encompasses elements from all previous quantisation solutions and can be represented with sparse formats, similar to weight pruning; and (ii) achieves very good inference accuracy.

\subsection{Data Formats Representation}

Neural network weights are commonly represented as matrices to support common algebraic operations. Since weight pruning and quantisation often leave the weight matrices very sparse (i.e.\ a high percentage of weights are zero-valued), sparse format representations are often used to reduce the matrix storage requirements. The most common sparse format used in large scale applications is Compressed Sparse Row (CSR) format\footnote{\url{http://www.scipy-lectures.org/advanced/scipy_sparse/csr_matrix.html}}. We employ this sparse representation for the machine learning techniques that introduce sparsity in the weight matrix (specifically, weight pruning and quantisation). We leave the exploration of other formats for future work.

%------------------------------------------------------------------------------------------------------------------------------------------------------

\begin{table*}[t]
\begin{center}
\caption{Experimental approach constructed on fundamental layers.}
\vspace{-0.1cm}
\begin{tabular}{ |c|c|c| } \hline
\bf{Experimental Layer} & \bf{Candidate Solutions} & \bf{Description} \\ \hline \hline
Neural Network Models & VGG-16, ResNet-18, MobileNet & CNNs trained on the CIFAR-10 dataset \\ \hline
Machine Learning Techniques & Deep Compression, Fisher pruning, Ternary Quantisation & Most common compression techniques \\ \hline
Data Formats and Algorithms & Dense/Sparse, \emph{im2col} & Network weights representation, matrix transformations \\ \hline
Systems Techniques & OpenMP/OpenCL, CLBlast & Parallel execution, libraries \\ \hline

Hardware & Odroid-XU4 (heterogeneous platform), Intel Core i7 & Resource-contained devices \\ \hline
\end{tabular}

\label{table:experiments_structure}
\end{center}
\end{table*}

\subsection{Systems Techniques}

We have developed parallel versions of each network model starting from serial versions we implemented in C. 
More specifically, each CNN is parallelised for CPU and GPU devices using OpenMP and OpenCL respectively. OpenMP 4.0 was used for the CPU parallel versions, whereas for the GPU versions we used OpenCL 1.1 and the OpenCL C++ wrapper API 1.1 which simplifies the host code. 

OpenMP is an API for shared memory parallelisation (i.e.\ all processors use the same address space). In our implementation, the outer \emph{for} loop of the convolutional layers is parallelised using dynamic scheduling of threads (because of the different amount of data required to process in each loop). Besides, the execution of the threads is synchronised on each neural network layer because each output is the input of the next layer, so we have to wait until all the operations from the previous layer finished. OpenMP suffers from some overheads such as threads initialisation and loops scheduling. 

In OpenCL programming, the main program which contains the specifications of the architecture and the workflow of a given network (host code) is executed on the CPU sequentially. Since the different layers of a network are parallelised and can potentially run on both CPU and GPU devices, the OpenCL code needs to be carefully designed to avoid data transfer overheads that may degrade the overall performance. OpenCL kernels communicate with the host code through buffers, which can be accessed directly from memory pointers. Therefore, the arrays in the GPU are handled as they are represented in memory, i.e.\ as 1-dimensional arrays. As a result, all the matrices are transformed to 1-dimensional arrays and passed through the buffers to the OpenCL kernels. This transformation is performed in the host code at the start of the program. Then, all layers handle 1-dimensional arrays and the final output is reformed back to a multidimensional matrix. Note that the transformation of the matrices is not a simple procedure, as it could lead to poor performance.

The specific optimisations used to obtain the parallel versions of the network models include threads (for both OpenMP and OpenCL), and work-groups, vectorisation (SIMD), and the CLBlast library \cite{Nugteren17} for OpenCL.

We use the CLBlast library \cite{Nugteren17} to transform the convolution operation to general matrix-matrix multiplication using the GEMM (Generalised Matrix - Matrix Multiplication) routine. It is well-known that matrix multiplication is one the of the most optimised operations in GPUs. However, using the CLBlast library for the convolution operation is not trivial, as it requires to change the structure of the matrices in a way that the matrix multiplication would give the same results as the convolution operation. This transformation can be done by using the \emph{im2col} operation, which rearranges image blocks to columns. Note that the CLBlast library can be tuned for a specific GPU architecture, and it includes an auto-tuner (CLTune) for that purpose. Up to 14 parameters can be tuned, for example: \emph{work group size}; \emph{register tiling configuration}; \emph{vector widths} of both matrices; \emph{loop unroll factors}; whether to use local memory or not; etc.

%------------------------------------------------------------------------------------------------------------------------------------------------------

\subsection{Hardware}

\subsubsection{Odroid-XU4} 

This board includes the ARM Cortex-A15 (4 cores @ 2.0GHz) and Cortex-A7 (4 cores @ 1.4GHz) big.LITTLE CPU, the low power ARM Mali T628 MP6 (6 shader cores @ 600MHz) GPU, and 2Gbyte of shared LPDDR3 RAM. The GPU supports 64-bit data types (scalar/vector, integer/float) which makes it suitable for accelerating applications that require significant computations like deep neural networks.

\subsubsection{Intel Core i7 CPU} A second platform we used for evaluation is a general purpose desktop system, with 4 cores Intel Core i7-3820 @ 3.60GHz, and 16GB DDR2. This was selected as a representative modern desktop system, which could be employed for inferences at the edge.

\subsection{Approach}

The experiments are conducted in a layered style, starting from the machine learning techniques (channel pruning, weight pruning, quantisation), moving down to data format representations (dense format, compressed sparse formats), followed by the systems layer where different parallelisation techniques are considered (OpenMP, OpenCL). We evaluate the networks on two target platforms representative of resource-constrained devices: an embedded heterogeneous system (Odroid-XU4) and a desktop CPU (Intel Core i7). Table Table~\ref{table:experiments_structure} presents a summary of the choices for each layer.

We chose to implement the three CNNs directly in C code for simplicity of control, allowing us to interchangeably use different libraries and to avoid the extra code contained in large deep learning frameworks necessary for other types of networks. We make our implementations available for the community to scrutinise and expand our work. We hope that this work can be used to direct decisions and priorities regarding future feature implementations in commonplace deep learning frameworks.

%---------------------------------------------------------------------------------------------------------------------------------------------

\begin{figure*}[t]
\includegraphics[width=\textwidth]{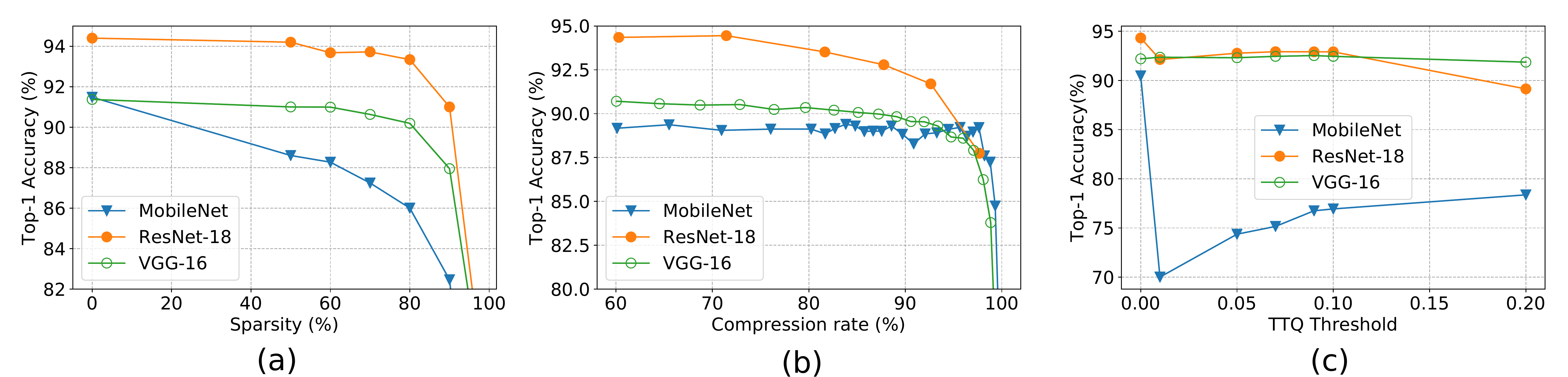}
\caption{Pareto curves of the accuracy and compression tradeoff at training time: (a) shows the accuracy of each model when weights are iteratively pruned to introduce an increasing level of sparsity into the weight matrices; (b) shows the rate of compression of the convolutional layers under the channel pruning scheme and the effect this has on error; (c) shows the accuracy of the models using an increasing quantisation threshold.}
\label{fig:accuracies}
\end{figure*}

\section{Experiments} 
\label{sec:experiments}

This section presents our results as we explore different candidates for each layer of the Deep Learning Inference Stack. We focus on three critical metrics: classification accuracy, inference time and memory footprint. We start by training our CNN models, and then apply three compression techniques separately to each model. Further down the stack, data formats are investigated by observing memory footprint and inference time on two target platforms (an Odroid-XU4 board and an Intel Core i7 machine), for (i) optimal accuracies, and (ii) fixed accuracies of 90\% for each model. Finally we explore the effect of parallelism by measuring the inference time of the more generic parallelisation libraries.

\subsection{Models Training}

Starting at the neural network model layer of the stack, we train each model from scratch with SGD using the hyper-parameters presented in Section~\ref{sec:hyperparameters} to obtain our baseline test accuracies. These are 92.20\%, 94.32\% and 90.47\% for VGG-16, ResNet-18 and MobileNet respectively.

\subsection{Compression Techniques Tuning}

\subsubsection{Weight Pruning}

We replicated Deep Compression~\cite{han2015deep} and for each model we set the initial threshold such that 50\% of weights (those with the lowest magnitude) are zeroed out. After fine-tuning the network for 30 epochs (a reasonable number of epochs for the models to converge) we increase the threshold and repeat to achieve greater sparsity in weights. 

The effect of this on the predictive performance of the models is outlined in Figure \ref{fig:accuracies}(a), which shows that both ResNet-18 and VGG-16 are able to withstand significant levels of weight pruning before losing accuracy. MobileNet, however, suffers significant accuracy losses which we attribute to its already optimised parameter count. Since this is a very aggressive method of parameter reduction, the network is not able to recover its original accuracy.

\subsubsection{Channel Pruning}

We implemented Fisher pruning, a channel-level pruning technique that removes whole channels, allowing the network to be recast as a new dense network with a reduced architecture, as opposed to the original architecture with increased sparsity. 

After initial training, the network is fine-tuned and a single channel is removed every 100 steps. A second-order Taylor expansion of the loss function is used to approximate the effect of removing each channel on the loss. In addition to this, a penalty is placed on each channel scaled by the number of floating point operations (FLOPs) it requires, meaning highly expensive channels are pruned first. The channel with the lowest effect on the loss, subject to a FLOP penalty, is removed. For our experiments, we fine-tune with a learning rate of \num{8e-3} for MobileNet and ResNet-18, and \num{8e-4} for VGG-16 and we use $\beta=\num{e-6}$ for the FLOP penalty.

The results of this are presented in Figure~\ref{fig:accuracies}(b). It is fascinating that all three networks perform very similarly as the compression rate increases. Note that ResNet-18 is not compressed as far as the other models for the network to remain in a standard form, as only layers between the shortcuts can be pruned.

\subsubsection{Quantisation}

Trained Ternary Quantisation (TTQ)~\cite{zhu_2017_trained} combines precision reduction and constraining weights to a small set of values for each layer. These values are determined iteratively over several epochs in two steps, first assessing the loss of output quality compared to a full precision version of the network and second adjusting the two values (one positive and one negative) such that loss of quality is minimised.

This method exposes a hyper-parameter (TTQ threshold) to control the separation between the weights, lower values being trimmed to zero. Beyond this threshold all weight values are forced to one determined positive value per layer and similarly for the negative values using the same threshold with negative sign. Intuitively this should control the sparsity level, although this is not necessarily implicit. As seen in Figure~\ref{fig:accuracies}(c), networks can converge to different weight distributions, which in the case of MobileNet is a flat distribution therefore requiring a larger threshold value to approximate those larger weights to single positive/negative values.

\subsection{Data format assignment}

Two of the compression techniques (weight pruning and quantisation) force many weights to zero. To avoid representing zero repeatedly in memory, the CSR format is used to store and to perform computations on for these two compression techniques. The storage space for the network weights is reduced considerably due to the large sparsity level (above 80\% for VGG-16 and ResNet-18 as seen in Figure~\ref{fig:accuracies}), although the memory footprint sees an increase (see Section~\ref{subsec:hardware}).  

Of course, weight-pruned models and quantised models can also be used in the dense format as done for the original models. However, in the rest of Section~\ref{sec:experiments} we use only the sparse format for these two compression techniques. Channel pruned models are evaluated in dense format.

\subsection{Baseline performance on the hardware platforms}
\label{subsec:hardware}

\begin{figure*}[ht]
	\includegraphics[width=\textwidth]{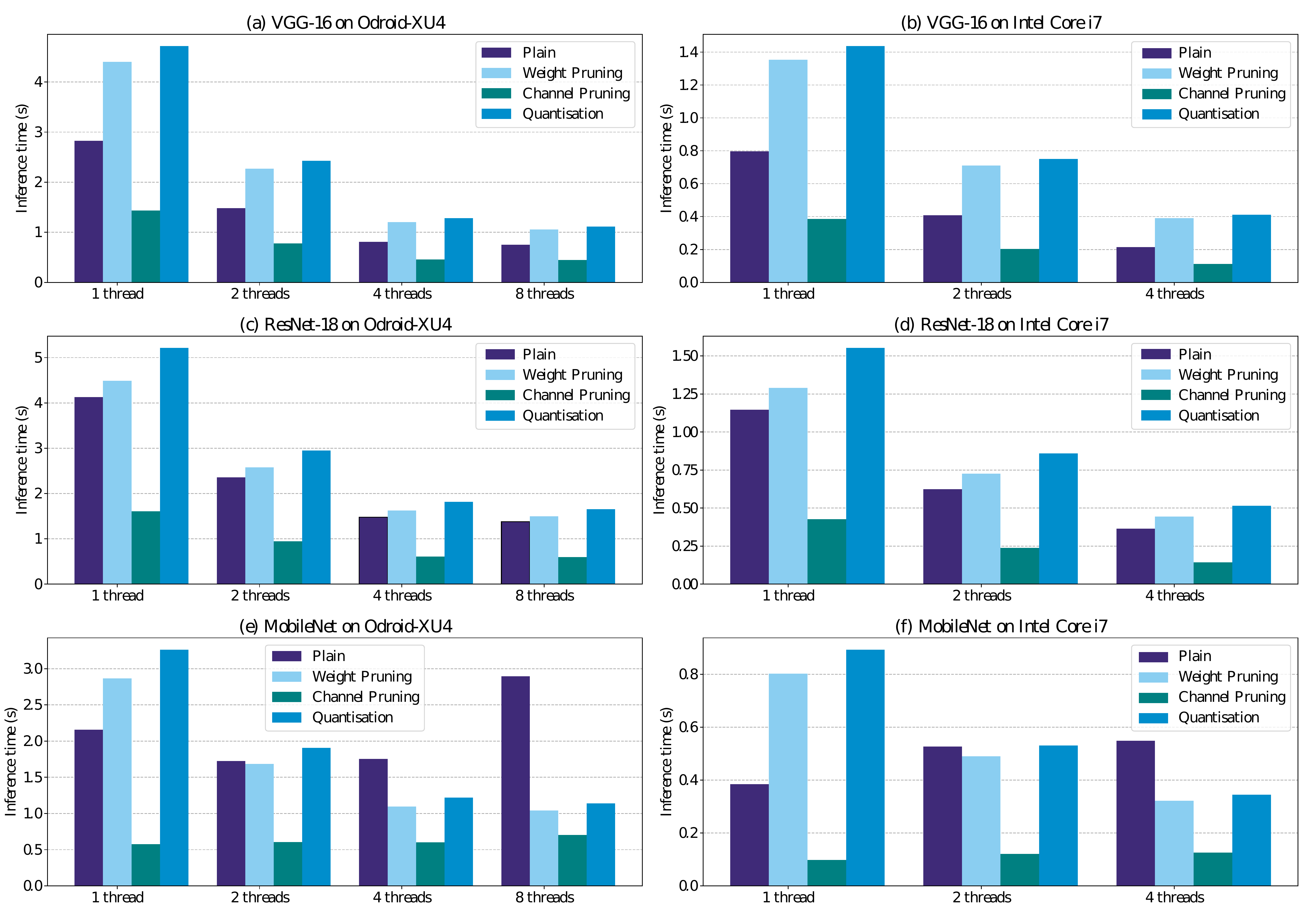}
    \caption{Baseline experiments comparing the compressed models chosen from obvious elbows on the Pareto curves of accuracy outlined in Table \ref{tab:pareto-points}, benchmarked on the Odroid-XU4 and Intel Core i7 platforms with an increasing thread count.}
  	\label{fig:baseline}
\end{figure*}

We now construct a full set of baselines by choosing an optimal point of accuracy on each Pareto curve (i.e.\ the elbow) in Figure~\ref{fig:accuracies} for each model. These optimal points are detailed in Table \ref{tab:pareto-points}, where \textit{sparsity} refers to the percentage of zero-valued parameters in the network, while \textit{compression} refers to the percentage of parameters that have been completely removed via channel pruning. Taking just one instance of each compression technique-model pair, we run our OpenMP implementations and show the inference time as thread count increases on both the Odroid-XU4 board and the Intel Core i7 CPU. These baseline results are reported in Figure~\ref{fig:baseline}. 

The key insight offered by the baseline experiments is that channel pruning significantly outperforms the other compression techniques in every setup considered. In some cases, the number of operations is larger in the channel-pruned model than the sparse format (for instance, the ResNet-18 models in Figure~\ref{fig:baseline}(d)) yet the inference time is still lower due to the other compression techniques using the sparse format representation (CSR). It is also worth noting that out of the three models, MobileNet is the least suitable for parallelisation, achieving no speedup on the two platforms (Figures~\ref{fig:baseline}(e)(f)); while for VGG-16 and ResNet-18 the inference times decrease when increasing the number of threads, we observe the opposite trend for MobileNet.

A second observation is that for VGG-16 and ResNet-18 the sparse methods (weight pruning and quantisation) fail to provide any speedup and do in fact hurt the performance of the network. Conversely, the inference time for the original model decreases when increasing the number of threads. However, MobileNet presents a different behaviour, as we see in Figures~\ref{fig:baseline}(e)(f), the sparse methods outperform the original model when increasing the number of threads.

Table~\ref{table:heap_allocation} presents the memory requirement at runtime for the original dense models and the three compression techniques. This is predominantly influenced by the network parameters being available in memory, input and output buffers and intermediate allocation for padding input in the convolutions. Again, channel pruning has an advantage of lower memory footprint due to fewer filters. In comparison with the original dense implementation, both weight pruning and quantisation have a slightly larger memory footprint despite being represented in CSR sparse format. This is due to using the direct convolution algorithm and the filter size of the networks, all relying on small $3\times3$ or $1\times1$ filters. Such dimensions require more memory space in sparse format than dense format: for example, in dense format the matrix is an array of 9 floating point elements for the $3\times3$ filter, while in CSR format there are 3 arrays holding the column offset, pointer to value on columns and the actual non-zero values, with additional parameters to account for the size of arrays. The memory footprint observation would be different for other algorithms implementation -- such as \textit{im2col}, which is not covered in these baseline experiments.

Through hashing at the level of bits, the memory requirement for quantisation could be an order of magnitude smaller although the inference time would also increase, which is the reason we chose not to compact the quantised format to its achievable minimum.

\begin{table}[t]
\caption{Compression rates for each model and compression technique for the baseline experiments.}
\begin{center}
\begin{tabular}{ |c|c|c|c| } \hline
& \thead{\textbf{W. Pruning} \\ Sparsity} & \thead{\textbf{C. Pruning} \\ Compres. Rate}     & \thead{\textbf{T. Quantisation} \\ TTQ thr. / Sparsity} \\ \hline \hline
\bf{VGG-16}    & 76.54\%  & 88.48\%  & 0.09 / 69.52\% \\ \hline
\bf{ResNet-18} & 88.92\%  & 60.24\%  & 0.07 / 87.93\% \\ \hline 
\bf{Mobilenet} & 23.46\%  & 80.33\%  & 0.20 / 92.13\% \\ \hline
\end{tabular}
\label{tab:pareto-points}
\end{center}
\end{table}

\begin{table}[t]
\begin{center}
\caption{Memory requirements (MB) for each model and compression technique for the baseline experiments.}
\begin{tabular}{ |c|c|c|c|c| } \hline
Model&Plain&W.~Pruning &C.~Pruning&T.~Quantis.\\ \hline \hline
VGG-16 &	111.4	& 144.4	& 17.9	& 130.3 \\ \hline
ResNet-18 &	89.0	& 99.4	& 31.6	& 100.8 \\ \hline
MobileNet & 69.1	& 188.5	& 10.8	& 201.1 \\ \hline
\end{tabular}
\label{table:heap_allocation}
\end{center}
\end{table}

\subsection{Accuracy Pareto Curve Exploration}

\begin{figure}[t]
\centering
\includegraphics[width=0.85\linewidth]{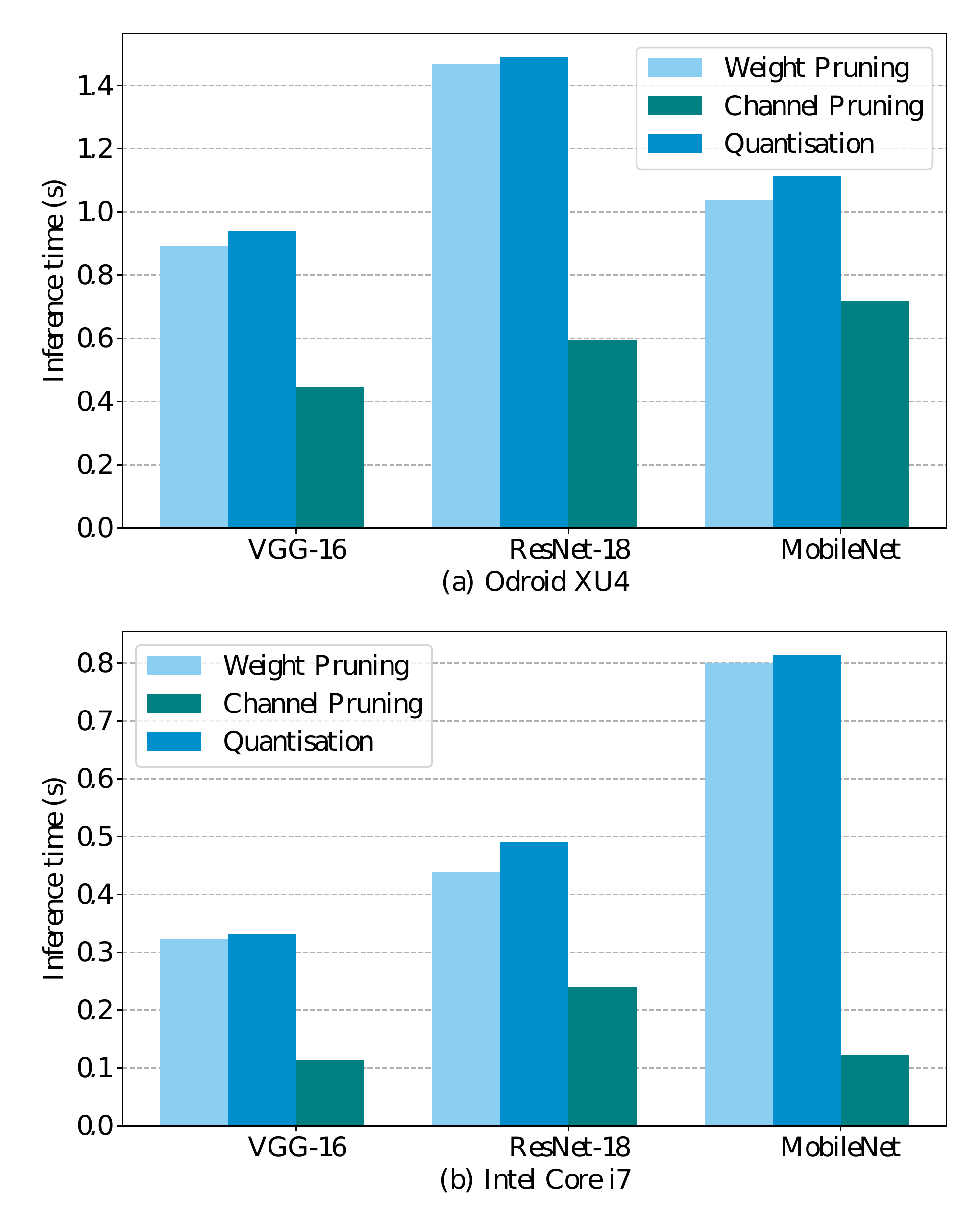}
\caption{Inference time comparison of weight pruning, channel pruning and quantisation on our three models when the accuracy level is fixed at 90\%: (a) Odroid-XU4 with eight threads; (b) Intel Core i7 with four threads.}
\label{fig:inf-mem}
\end{figure}

In the previous section we compared the elbows of the Pareto curves for each model to highlight which compression technique gives the best performance in terms of accuracy. This is a useful comparison when accuracy is a high priority, however, in some scenarios we may be willing to accept a larger loss in accuracy for increased performance on hardware. In this section we show an example of how this tradeoff can affect model choice by constraining the accuracy of all of the compressed models to 90\%; this puts the focus more heavily on the hardware performance of the technique, rather than the accuracy achievable by the model. We chose 90\% as it is achievable by all of the networks without demanding a significant drop in accuracy. Table~\ref{tab:pareto-points-90} shows the compression rates for the chosen 90\% accuracy level. Using these values, Figure~\ref{fig:inf-mem} and Table~\ref{table:mem-90} show the inference time and memory footprint of the three networks respectively.

On both hardware platforms the speedup gained from channel pruning is clear. Furthermore, the memory footprint of these networks is reduced significantly beyond the memory reduction achieved by the sparse methods. However, it is interesting to note that on the Odroid board the inference time of MobileNet is slower than both ResNet-18 and VGG-16 (Figure \ref{fig:inf-mem}). With accuracy fixed at 90\%, we are able to use channel pruning to optimise very large networks (VGG-16 and ResNet-18) and outperform a small network hand-tuned specifically for embedded inference (MobileNet).

\begin{table}[h]
\caption{Compression rates for each model and compression technique when accuracy is fixed at 90\%.}
\begin{center}
\begin{tabular}{ |c|c|c|c| } \hline
& \thead{\textbf{W. Pruning} \\ Sparsity} & \thead{\textbf{C. Pruning} \\ Compres. Rate}     & \thead{\textbf{T. Quantisation} \\ TTQ thr. / Sparsity} \\ \hline \hline
\bf{VGG-16}    & 85.00\%  & 94.00\%  & 0.2 / 70.00\% \\ \hline %0.09 \\ \hline
\bf{ResNet-18} & 91.00\%  & 94.00\%  & 0.2 / 80.00\% \\ \hline %0.07 \\ \hline
\bf{Mobilenet} & 42.00\%  & 96.00\%  & 0.2 / 20.00\% \\ \hline %0.20 \\ \hline
\end{tabular}
%\vspace{-0.2cm}
\label{tab:pareto-points-90}
\end{center}
\end{table}

\begin{table}[h]
\begin{center}
\caption{Memory requirements (MB) for each model and compression technique when accuracy is fixed at 90\%.}
%\vspace{-0.2cm}
\begin{tabular}{ |c|c|c|c|c| } \hline
Model&Plain&W. Pruning &C.~Pruning&T. Quantis.\\ \hline \hline
VGG-16 	  &	309.9	& 112.2	& 74.9	& 114.1 \\ \hline
ResNet-18 &	233.8	& 66.1	& 13.1	& 66.9 \\ \hline
MobileNet & 66.3	& 40.9	& 2.7	& 63.3 \\ \hline
\end{tabular}
%\vspace{-0.8cm}
\label{table:mem-90}
\end{center}
\end{table}

Although we explore only the accuracy Pareto Curve of the models and take memory footprint and inference time as observable parameters for the compression techniques in question, it is relevant to explore further by fixing specific memory requirements or inference times. We leave this exploration for future work.

\subsection{Parallelisation in heterogeneous systems}

We also consider different parallel implementations, using OpenMP and OpenCL, for the dense models of the three networks when running on the Odroid-XU4 board. 

Since OpenMP does not support ARM Mali GPUs, the networks are parallelised only on the CPU of the board using up to 8 threads (cores) of the Cortex-A processor. 

We developed two OpenCL versions. The first one uses the CLBLast library to perform the convolution operation as a general matrix multiplication. The second one performs dot products for the convolutions and we hand-tuned the parameters in order to get the best performance. Specifically we chose $4 \times 4$ work-items for the work-group size and we vectorised the code using vectors of 16 elements.

%~\ref{fig:parallel}
Figure 6 shows the results for the inference time when comparing the three parallel versions of each network. As we see, the hand-tuned OpenCL versions outperform the OpenMP implementations. The parameters selected perform very well when mapping the OpenCL kernels on the GPU cores. This shows that GPUs are preferable even on embedded devices to CPUs for neural network workloads. However, the version using the CLBlast library hurt performance, suffering up to a $10\times$ slowdown on ResNet-18. This poor performance is due to the small size of CIFAR-10 input images ($32\times32$ pixels); the efficient matrix multiplication operation only pays off for big matrices~\cite{multiprog_2018} so the bigger the matrix the greater the improvement. As such, when using the ImageNet dataset for VGG-16 (where images are $224\times224$ pixels) the CLBlast library actually outperforms the OpenMP implementations.

\begin{figure}
\centering
\includegraphics[width=0.85\linewidth]{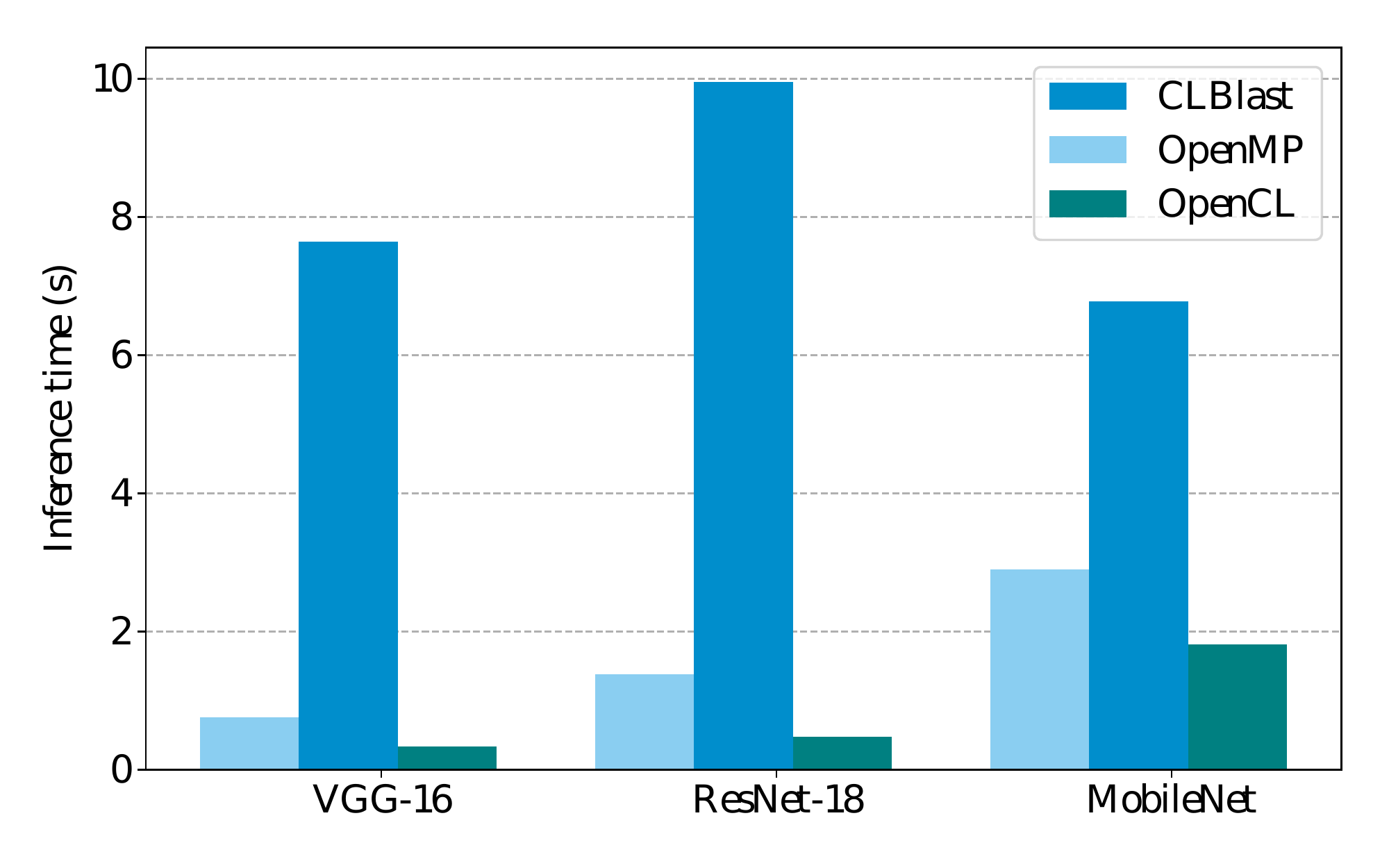}
\label{fig:parallel}
\caption {Performance of the plain models when parallelised using OpenMP (with 8 threads), CLBlast and hand-tuned OpenCL on the Odroid-XU4 board.}
\end{figure}

%% file: conclusions.tex
\section{Discussion}
\label{sec:discussion}

Looking across the Deep Learning Inference Stack we can make some very interesting observations. Some support existing neural network lore, while others are contradictory to commonly held beliefs.

Our memory footprint analysis validates that memory access is a bottleneck for neural network computations. In fact, this is more obvious with the sparse formats where memory continuity is not achieved due to many arrays being allocated to represent and access only non-zero values. In the case of CNNs with small filters ($3\times3$ convolutions), the sparse format actually takes up more memory. For this reason introducing sparsity in the convolutional layers would only truly benefit networks with larger filter sizes. This is a contradictory observation to the general belief in machine learning that sparsity can be useful for adapting neural network models to run more efficiently on smaller devices. In the code optimisation community, the knowledge that 3$\times$3 convolutions are the current standard in neural network architectures would allow for the creation of more efficient data structures, even though these would not readily generalise to other filter sizes.

Although the CLBlast library is known to be very efficient at performing basic linear algebra operations --- often regarded as the most promising candidate to parallelise the operations in neural networks --- we found that in many cases a hand-coded implementation with OpenCL and OpenMP is faster. This shows that overheads imposed by specialised libraries surpass their benefits in highly optimised matrix operations, which is important for deep learning framework designers.

Collectively, these observations expose a gap between the machine learning community --- proposing solutions that are not beneficial to hardware execution --- and the systems community that proposes libraries for executing neural networks without a full understanding of the workloads involved. Fundamentally, to optimise a deep learning model to run on a particular device requires careful selection and tuning of techniques across the Deep Learning Inference Stack and this can be achieved only by the two research communities working together. 

\section{Conclusions}

This paper presents a characterisation of current state-of-the-art solutions across the Deep Learning Inference Stack and their limitations. We find that promising candidate solutions to compress Convolutional Neural Networks (i.e.\ weight pruning and quantisation) hurt performance given libraries and hardware that are not able to leverage the reduction in parameter count into real speedup. It is critical to overcome these limitations in order to bring neural networks to the edge. We believe this can be achieved through closer collaboration across the layers of the stack; to produce machine learning techniques and models with a view of deployment on the resource-contained hardware.